\documentclass[10pt,twocolumn,letterpaper]{article}

\usepackage[pagenumbers]{wacv} 

\usepackage{graphicx}
\usepackage{amsmath}
\usepackage{amssymb}
\usepackage{booktabs}

\usepackage{listings}
\usepackage{xcolor}
\definecolor{codegreen}{rgb}{0,0.6,0}
\definecolor{codegray}{rgb}{0.5,0.5,0.5}
\definecolor{codepurple}{rgb}{0.58,0,0.82}
\definecolor{backcolour}{rgb}{0.95,0.95,0.92}

\definecolor{Red}{RGB}{255, 153, 151}
\definecolor{LightRed}{RGB}{255, 224, 224}

\definecolor{Green}{RGB}{210, 253, 187}
\definecolor{Greener}{RGB}{210, 253, 210}
\definecolor{LightGreen}{RGB}{236, 255, 239}

\definecolor{Orange}{RGB}{255,165,0}
\definecolor{Aqua}{RGB}{50, 235, 235}
\definecolor{Purple}{RGB}{150, 150, 255}
\definecolor{Blue}{RGB}{174, 198, 255}

\definecolor{Yellow}{RGB}{249, 249, 200}
\definecolor{Mega}{RGB}{229, 229, 253}
\definecolor{ImageNet}{RGB}{231, 242, 230}
\definecolor{DINO}{RGB}{247, 207, 224}

\lstdefinestyle{mystyle}{
    backgroundcolor=\color{backcolour},   
    commentstyle=\color{codegreen},
    keywordstyle=\color{magenta},
    numberstyle=\tiny\color{codegray},
    stringstyle=\color{codepurple},
    basicstyle=\ttfamily\footnotesize,
    xleftmargin=10pt,
    breakatwhitespace=false,         
    breaklines=true,                 
    captionpos=b,                    
    keepspaces=true,                 
    numbers=left,                    
    numbersep=5pt,                  
    showspaces=false,                
    showstringspaces=false,
    showtabs=false,                  
    tabsize=2
}

\usepackage{pifont}
\newcommand{\cmark}{\ding{51}}%
\newcommand{\xmark}{\ding{55}}%

\usepackage[pagebackref,breaklinks,colorlinks]{hyperref}

\usepackage[capitalize]{cleveref}
\crefname{section}{Sec.}{Secs.}
\Crefname{section}{Section}{Sections}
\Crefname{table}{Table}{Tables}
\crefname{table}{Tab.}{Tabs.}

\def\wacvPaperID{1249} 
\def\confName{WACV}
\def\confYear{2024}

\newcommand\blfootnote[1]{%
  \begingroup
  \renewcommand\thefootnote{}\footnote{#1}%
  \addtocounter{footnote}{-1}%
  \endgroup
}

\begin{document}

\title{WildlifeDatasets: An open-source toolkit for animal re-identification}

\author{Vojtěch Čermák$^1$, Lukas Picek$^{2,3}$, Lukáš Adam$^1$ and Kostas Papafitsoros$^4$\\
$^1$Czech Technical University in Prague, $^2$University of West Bohemia, $^3$INRIA\\ $^4$Queen Mary University of London \\
{\tt\small cermavo3@fel.cvut.cz}, {\tt\small picekl@kky.zcu.cz/lpicek@inria.cz} \\{\tt\small lukas.adam.cr@gmail.com}, {\tt\small k.papafitsoros@qmul.ac.uk}
}
\maketitle

\begin{abstract}
In this paper, we present \href{https://github.com/WildlifeDatasets/wildlife-datasets}{WildlifeDatasets} -- an open-source toolkit
intended primarily for ecologists and computer-vision / machine-learning researchers. The WildlifeDatasets is written in Python, allows straightforward access to publicly available wildlife datasets, and provides a wide variety of methods for dataset pre-processing, performance analysis, and model fine-tuning.
We showcase the toolkit in various scenarios and baseline experiments, including, to the best of our knowledge, the most comprehensive experimental comparison of datasets and methods for wildlife re-identification, including both local descriptors and deep learning approaches.
Furthermore, we provide the first-ever foundation model for individual re-identification within a wide range of species -- MegaDescriptor -- that provides state-of-the-art performance on animal re-identification datasets and outperforms other pre-trained models such as CLIP and DINOv2 by a significant margin. To make the model available to the general public and to allow easy integration with any existing wildlife monitoring applications, we provide multiple MegaDescriptor flavors (i.e., Small, Medium, and Large) through the \href{https://huggingface.co/BVRA}{HuggingFace hub}.

\end{abstract}

\section{Introduction}
\label{sec:intro}

Animal re-identification is essential for studying different aspects of wildlife, like population monitoring, movements, behavioral studies, and wildlife management \cite{papafitsoros_2021,Schofield_2022,vidal2021perspectives}. While the precise definition and approaches to animal re-identification may vary in the literature, the objective remains consistent. The main goal is to accurately and efficiently recognize individual animals within one species based on their unique characteristics, e.g., markings, patterns, or other distinctive features.

Automatizing the identification and tracking of individual animals enables the collection of precise and extensive data on population dynamics, migration patterns, habitat usage, and behavior, facilitating researchers in monitoring movements, evaluating population sizes, and observing demographic shifts. This invaluable information contributes to a deeper comprehension of species dynamics, identifying biodiversity threats, and developing conservation strategies grounded in evidence.

\begin{figure}[!t]
\vspace{0.25cm}
\centering
MegaDescriptor~~ \hspace{2.0cm} DINOv2~~~~~~ \\
        \includegraphics[width=0.475\linewidth]{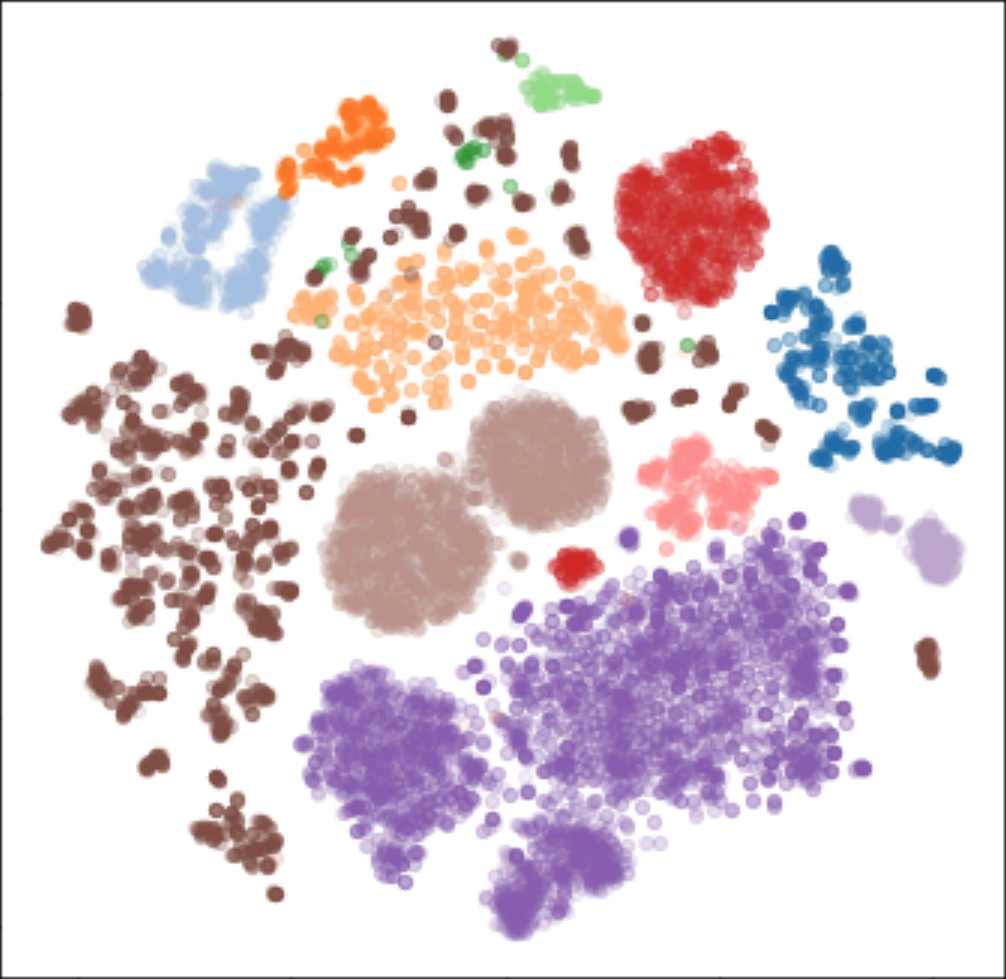}
        \includegraphics[width=0.475\linewidth]{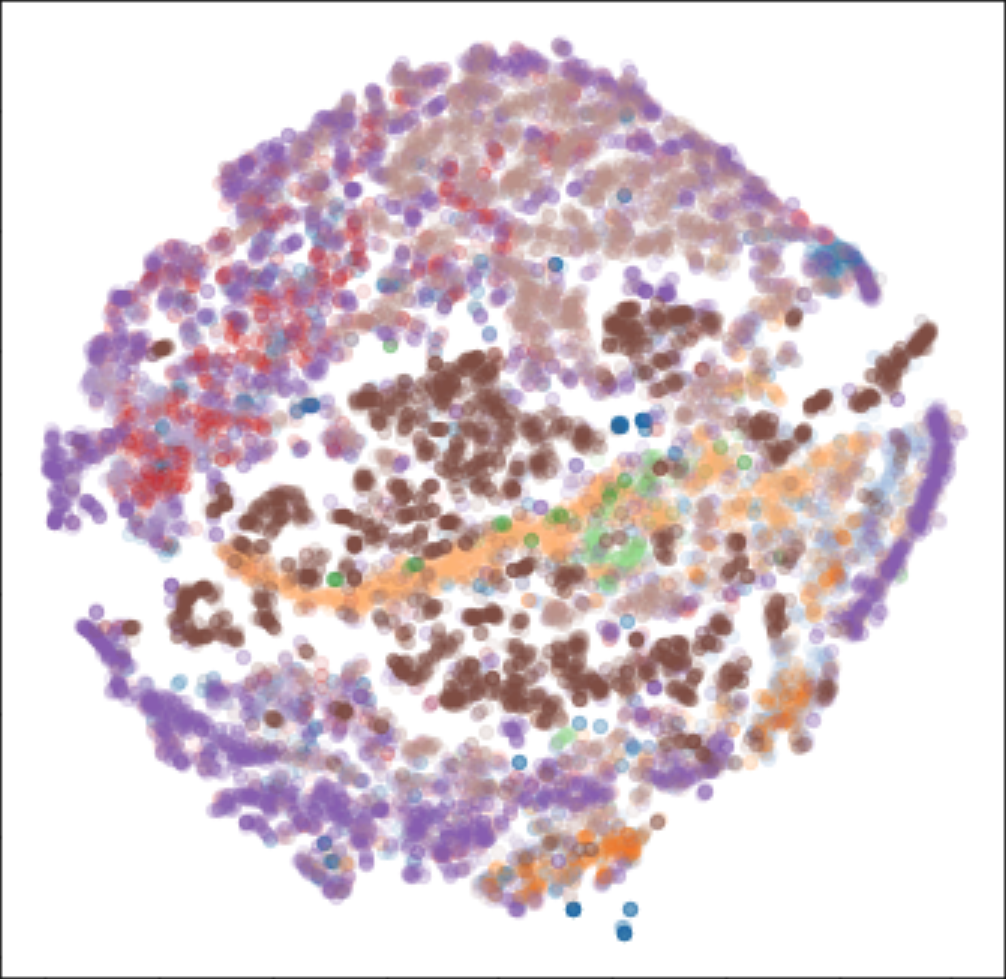}
\caption{\textbf{Latent space separability of MegaDescriptor}. Embedding visualization (t-sne) of unseen individual animals (identity-wise) for the proposed  MegaDescriptor and DINOv2. Colors represent different datasets (i.e., species).}
\label{fig:embeddings}
\vspace{-0.25cm}
\end{figure}

Similarly, the increasing sizes of the collected data and the increasing demand for manual (i.e., time-consuming) processing of the data highlighted the need for automated methods to reduce labor-intensive human supervision in individual animal identification. As a result, a large number of automatic re-identification datasets and methods have been developed, covering several animal groups like primates\,\,\cite{freytag2016chimpanzee,witham2018automated}, carnivores\,\,\cite{li2019atrw,botswana2022,dlamini2020automated}, reptiles\,\,\cite{zinditurtles, dunbar2021hotspotter}, whales\,\,\cite{humpbackwhale,rightwhale,cheeseman2017happywhale}, and mammals\,\,\cite{belugaid,trotter2020ndd20,zuffi2019three}. 

However, there is a lack of standardization in algorithmic procedures, evaluation metrics, and dataset utilization across the literature. This hampers the comparability and reproducibility of results, hindering the progress of the field. It is, therefore, essential to categorize and re-evaluate general re-identification approaches, connect them to real-world scenarios, and provide recommendations for appropriate algorithmic setups in specific contexts. By quantitatively assessing the approaches employed in various studies, we aim to identify trends and provide insights into the most effective techniques for different scenarios. This analysis will aid researchers and practitioners in selecting suitable algorithms for their specific re-identification needs, ultimately advancing the field of animal re-identification and its applications in wildlife conservation and research.

To address these issues, we have developed an open-source toolkit -- \href{https://github.com/WildlifeDatasets/wildlife-datasets}{WildlifeDatasets} -- intended primarily for ecologists and computer-vision / machine-learning researchers. 
In this paper, besides the short description of the main features of our tool, 
(i) we list all publicly available wildlife re-identification datasets,
(ii) perform the largest experimental comparison of datasets and wildlife re-identification methods,
(iii) describe a foundation model -- MegaDescriptor -- based on different Swin architectures and trained on a newly comprised dataset, and
(iv) provide a variety of pre-trained models on a \href{https://huggingface.co/BVRA}{HuggingFace hub}.

\section{Related work}

Similarly, as in other fields, the development of methods and datasets for automated animal re-identification has been influenced by the progress in machine learning. Currently, many studies exist, although the differences in terms of their approach, prediction output, and evaluation methodologies result in several drawbacks.

\textit{Firstly}, methods are usually inspired by trends in machine learning rather than being motivated by real-world re-identification scenarios. A prominent example is performing classification tasks on a closed-set, which is typical for benchmarking in deep learning but is, in general, not realistic in ecology, as new individuals are constantly being recruited to populations.

\textit{Second}, many studies focus on a single dataset and develop species-specific methods evaluated on the given dataset rather than on a family of datasets\cite{bedetti2020system,weideman2020extracting,gilman2016computer,li2019atrw,drechsler2015genetic,anderson2010computer}, making reproducibility, transferability, and generalization challenging. 

\textit{Third}, datasets are poorly curated and usually include unwanted training-to-test data leakage, which leads to inflated performance expectations. 

All this leads to the repetition of poor practices both in dataset curation and method design. As such, much of the current research suffers from a lack of unification, which, we argue, constitutes an obstacle to further development, evaluation, and applications to real-world situations.

\subsection{Tools and methods}

There are three primary approaches commonly used for wildlife re-identification -- (i) local descriptors \cite{reno2019sift,andrew2016automatic,dunbar2021hotspotter}, (ii) deep descriptors \cite{bruslund2020re,ueno2022automatic,li2019atrw,deb2018face,miele2021revisiting}, and (iii) species-specific methods \cite{bedetti2020system,weideman2020extracting,gilman2016computer,kelly2001computer,anderson2010computer}. \\

\newpage

\noindent\textbf{Local-feature-based methods} find unique keypoints and extract their local descriptors for matching. The matching is usually done on a database of known identities, i.e., for each given image sample, an identity with the highest number of descriptor matches is retrieved.
The most significant benefit of these methods is their plug-and-play nature, without any need for fine-tuning, which makes them comparable in a zero-shot setting to large foundation models, such as CLIP \cite{radford2021learning} or DINOv2 \cite{oquab2023dinov2}, etc. 

Even though approaches based on SIFT, SURF, or ORB descriptors exhibit limitations in scaling efficiently to larger datasets and their performance, all available software products, e.g., WildID \cite{bolger2012computer}, HotSpotter \cite{crall2013hotspotter}, and \href{https://github.com/daniel-brenot/I3S-Interactive-Individual-Identification-System-Desktop}{I$^3$S}, are based on local-feature-based methods. 
Naturally, even with such limitations, those systems are popular among ecological researchers without a comprehensive technical background and find a wide range of applications, most likely due to their intuitive graphical user interfaces (GUIs). \\

\textbf{Deep feature-based approaches} are based on vector representation of the image learned through optimizing a deep neural network. Similarly, as in local feature-based methods, the resulting deep embedding vector (usually 1024 or 2048d) is matched with an identity database.

Applying deep learning to wildlife re-identification bears similarities with human or vehicle re-identification. Therefore, similar methods can be easily repurposed. However, it is important to note that deep learning requires fine-tuning models on the specific target domain, i.e., species, which makes the model's performance dependent on a species it was fine-tuned for. 
Another approach is to use publicly available large-scale, foundational models pre-trained on large datasets (e.g., CLIP \cite{radford2021learning} and DINOv2 \cite{oquab2023dinov2}). These models are primarily designed for general computer vision tasks. Therefore, they are not adapted nor tested for the nuances of wildlife re-identification, which heavily relies on fine-grained features. \\

\noindent\textbf{Species-specific methods} are tailored to an individual species or groups of closely related species, particularly those with visually distinct patterns. These methods typically focus on visual characteristics unique to the target species, restricting their applicability beyond the species they were developed for. Moreover, they often entail substantial manual preprocessing steps, such as extracting patches from regions of interest or accurately aligning compared images. For instance, one such approach involves employing Chamfer distance to measure the distance between greyscale patterns in polar bear whiskers \cite{anderson2010computer}. Other examples include computing correlation between aligned patches derived from cheetah spots \cite{kelly2001computer} or similarity between two images based on the count of matching pixels within newt patterns \cite{drechsler2015genetic}.

\newpage
\section{The WildlifeDatasets toolkit}

One of the current challenges for the advancement of wildlife re-identification methods is the fact that datasets are scattered across the literature and that adopted settings and developed algorithms heavily focus on the species of interest. In order to facilitate the development and testing of re-identification methods across multiple species in scale and evaluate them in a standardized way, we have developed the Wildlife Datasets toolkit consisting of two Python libraries -- \href{https://github.com/WildlifeDatasets/wildlife-datasets}{WildlifeDatasets} and \href{https://github.com/WildlifeDatasets/wildlife-tools}{WildlifeTools}\footnote{Both libraries are available online on \href{https://github.com/WildlifeDatasets/wildlife-datasets}{ GitHub}.}.
Both libraries are \href{https://wildlifedatasets.github.io/wildlife-datasets}{documented} in a user-friendly way; therefore, it is accessible to both animal ecologists and computer vision experts. Users just have to provide the data and select the algorithm. Everything else can be done using the toolkit: extracting and loading data, dataset splitting, identity matching, evaluation, and performance comparisons. Experiments can be done over one or multiple datasets fitting into any used specified category, e.g., size, domain, species, and capturing conditions. Below, we briefly describe the core features and use cases of both libraries.

\begin{table}[!hb]
\vspace{-0.5cm}
\small
\setlength{\tabcolsep}{0.3em} 
\centering
\begin{tabular}{@{}lcrr@{\hskip 4mm}l@{\hskip 2mm}l@{\hskip 2mm}l@{\hskip 2mm}l@{}}
\toprule
                                            \textbf{Name} &  \textbf{Year} & \# \textbf{Images} & \rotatebox{90}{\# \textbf{\footnotesize{Identities}}}  & \textbf{\rotatebox{90}{\footnotesize{Timestamp}}} & \textbf{\rotatebox{90}{\footnotesize{In-the-wild}}} & \textbf{\rotatebox{90}{\footnotesize{Pattern}}} & \textbf{\rotatebox{90}{\footnotesize{Multispecies}}} \\
\midrule
            AAUZebraFishID \cite{bruslund2020re} &  2020 &          6672 &             6 &                         \xmark &               \xmark &                  \xmark &                           \xmark \\
        AerialCattle2017 \cite{andrew2017visual} &  2017 &         46340 &            23 &                        \xmark &               \xmark &                  \cmark &                           \xmark \\
                          ATRW \cite{li2019atrw} &  2019 &          5415 &           182 &                         \xmark &               \xmark &                  \cmark &                           \xmark \\
                        BelugaID \cite{belugaid} &  2022 &          5902 &           788 &                      \cmark &               \cmark &                  \xmark &                           \xmark \\
        BirdIndividualID \cite{ferreira2020deep} &  2019 &         51934 &            50 &                        \xmark &               \xmark &                  \xmark &                           \cmark \\
               CTai \cite{freytag2016chimpanzee} &  2016 &          4662 &            71 &                        \xmark &               \cmark &                  \xmark &                           \xmark \\
               CZoo \cite{freytag2016chimpanzee} &  2016 &          2109 &            24 &                         \xmark &               \xmark &                  \xmark &                           \xmark \\
                  Cows2021 \cite{gao2021towards} &  2021 &          8670 &           181 &                       \cmark &               \xmark &                  \cmark &                           \xmark \\
              Drosophila \cite{schneider2018can} &  2018 &          $\sim$2.6M &            60 &                      \xmark &               \xmark &                  \cmark &                           \xmark \\
   FriesianCattle2015 \cite{andrew2016automatic} &  2016 &           377 &            40 &                       \xmark &               \xmark &                  \cmark &                           \xmark \\
      FriesianCattle2017 \cite{andrew2017visual} &  2017 &           940 &            89 &                         \xmark &               \xmark &                  \cmark &                           \xmark \\
          GiraffeZebraID \cite{parham2017animal} &  2017 &          6925 &          2056 &                      \cmark &               \cmark &                  \cmark &                           \cmark \\
             Giraffes \cite{miele2021revisiting} &  2021 &          1393 &           178 &                        \xmark &               \cmark &                  \cmark &                           \xmark \\
       HappyWhale \cite{cheeseman2017happywhale} &  2022 &         51033 &         15587 &                   \xmark &               \cmark &                  \cmark &                           \xmark \\
            HumpbackWhaleID \cite{humpbackwhale} &  2019 &         15697 &          5004 &                     \xmark &               \cmark &                  \cmark &    
            \xmark \\
                 HyenaID2022 \cite{botswana2022} &  2022 &          3129 &           256 &                         \xmark &               \cmark &                  \cmark &                           \xmark \\
                   IPanda50 \cite{wang2021giant} &  2021 &          6874 &            50 &                        \xmark &               \xmark &                  \cmark &                           \xmark \\
               LeopardID2022 \cite{botswana2022} &  2022 &          6806 &           430 &                         \xmark &               \cmark &                  \cmark &                           \xmark \\
            LionData \cite{dlamini2020automated} &  2020 &           750 &            94 &                       \xmark &               \cmark &                  \cmark &                           \xmark \\
         MacaqueFaces \cite{witham2018automated} &  2018 &          6280 &            34 &                       \cmark &               \xmark &                  \xmark &                           \xmark \\
                   NDD20 \cite{trotter2020ndd20} &  2020 &          2657 &            82 &                        \xmark &               \xmark &                  \cmark &                           \xmark \\
                NOAARightWhale \cite{rightwhale} &  2015 &          4544 &           447 &                        \xmark &               \cmark &                  \xmark &                           \xmark \\
           NyalaData \cite{dlamini2020automated} &  2020 &          1942 &           237 &                       \xmark &               \cmark &                  \cmark &                           \xmark \\
            OpenCows2020 \cite{andrew2021visual} &  2020 &          4736 &            46 &                         \xmark &               \xmark &                  \cmark &                           \xmark \\
             SealID \cite{nepovinnykh2022sealid} &  2022 &          2080 &            57 &                      \xmark &               \cmark &                  \cmark &                           \xmark \\
  SeaTurtleID \cite{papafitsoros2022seaturtleid} &  2022 &          7774 &           400 &                       \cmark &               \cmark &                  \cmark &                           \xmark \\
    SeaTurtleID2022 \cite{SeaTurtleID2022} &  2024 &      8729     &      438      &                       \cmark &               \cmark &                  \cmark &                           \xmark \\

                    SMALST \cite{zuffi2019three} &  2019 &         12850 &            10 &                     \xmark &               \xmark &                  \cmark &                           \xmark \\
        StripeSpotter \cite{lahiri2011biometric} &  2011 &           820 &            45 &                       \xmark &               \cmark &                  \cmark &                           \xmark \\
      WhaleSharkID \cite{holmberg2009estimating} &  2020 &          7693 &           543 &                     \cmark &               \cmark &                  \cmark &                           \xmark \\
           ZindiTurtleRecall \cite{zinditurtles} &  2022 &         12803 &          2265 &                       \xmark &               \cmark &                  \cmark &                           \xmark \\
\bottomrule
\end{tabular}
\caption{\textbf{Publicly available animal re-identification datasets}. We list all datasets for animal re-identification and their relevant statistics, e.g., number of images, identities, etc. All listed datasets are available for download in the WildlifeDatasets toolkit.}
\label{table:datasets}
\end{table}

\subsection{All publicly available wildlife datasets at hand}
The first core feature of the WildlifeDatasets toolkit allows downloading, extracting, and pre-processing all 31 publicly available wildlife datasets\footnote{Based on our research at the end of September 2023.} (refer to Table\,\ref{table:datasets}) in a unified format using just a few lines of Python code. For reference, see provided code snippet in Figure\,\ref{fig:downloading}. 
Additionally, users can quickly overview and compare images of the different datasets and their associated metadata, e.g., image samples, number of identities, timestamp information, presence of segmentation masks/bounding boxes, and general statistics about the datasets. This feature decreases the time necessary for data gathering and pre-processing tremendously.
Recognizing the continuous development of the field, we also provide user-friendly options for adding new datasets.

\subsection{Implementation of advanced dataset spliting}

Apart from the datasets at hand, the toolkit has built-in implementations for all dataset training/validation/test splits corresponding to the different settings, including (i) \textit{closed-set} with the same identities in training and testing sets, (ii) \textit{open-set} with a fraction of newly introduced identities in testing, and (iii) \textit{disjoint-set} with different identities in training and testing. In cases where a dataset contains timestamps, we provide so-called time-aware splits where images from the same period are all in either the training or the test set. This results in a more ecologically realistic split where new factors, e.g., individuals and locations, are encountered in the future \cite{papafitsoros2022seaturtleid}.

\begin{figure}[!h]
    \vspace{0.25cm}
    \centering
\begin{lstlisting}[language=Python,linewidth=1.0\linewidth]
#Import wildlife-datasets Library
from wildlife_datasets import datasets, splits

#Download dataset
datasets.ATRW.get_data('data/ATRW')

#Load metadata
metadata = datasets.ATRW('data/ATRW')

#Get 80/20 training/test split
splitter = splits.ClosedSetSplit(0.8)

splitter.split(metadata.df)
>>> [<train indices>, <test indices>]
\end{lstlisting}
    \vspace{-0.25cm}
    \caption{\textbf{Dataset download with WildlifeDatasets}. A code snippet showcasing easy data download, metadata load, and splitting.}
    \label{fig:downloading}
    \vspace{-0.5cm}
\end{figure}

\newpage

\subsection{Accessible feature extraction and matching}

Apart from the datasets, the WildlifeDatasets toolkit provides the ability to access multiple feature extraction and matching algorithms easily and to perform re-identification on the spot. We provide a variety of local descriptors, pre-trained CNN- and transformer-based descriptors, and different flavors of the newly proposed foundation model -- MegaDescriptor. Below, we provide a short description of all available methods and models.   \vspace{-0.1cm}\\

\noindent\textbf{Local descriptors}: Due to extensive utilization among ecologists and state-of-the-art performance in animal re-identification, we have included selected local feature-based descriptors as a baseline solution available for deployment and a direct comparison with other approaches. 

Within the toolkit, we have integrated our implementations of standard SIFT and deep learning-based Superpoint descriptors. Besides, we have implemented a matching algorithm that uses local descriptors using contemporary insights and knowledge. Leveraging GPU implementation (FAISS\cite{johnson2019billion}) for nearest neighbor search, we have eliminated the necessity for using approximate neighbors. This alleviates the time-complexity concerns raised by authors of the Hotspotter tool.  \vspace{-0.1cm}\\

\noindent\textbf{Pre-trained deep-descriptors}: Besides local descriptors, the toolkit allows to load any pre-trained model available on the HuggingFace hub and to perform feature extraction over any re-identification datasets. We have accomplished this by integrating the Timm library\cite{rw2019timm}, which includes state-of-the-art CNN- and transformer-based architectures, e.g., ConvNeXt\,\cite{liu2022convnet}, ResNext\,\cite{next}, ViT\,\cite{dosovitskiy2020image}, and Swin\,\cite{liu2021swin}. This integration enables both the feature extraction and the fine-tuning of models on the wildlife re-identification datasets. \vspace{-0.1cm}\\

\noindent\textbf{MegaDescriptor}: Furthermore, we provide the first-ever foundation model for individual re-identification within a wide range of species -- MegaDescriptor -- that provides state-of-the-art performance on all datasets and outperforms other pre-trained models such as CLIP and DINOv2 by a significant margin. In order to provide the models to the general public and to allow easy integration with any existing wildlife monitoring applications, we provide multiple MegaDescriptor flavors, e.g., Small, Medium, and Large, see Figure\,\,\ref{fig:feature_prediction} for reference. \vspace{-0.1cm} \\

\noindent\textbf{Matching}:
Next, we provide a user-friendly high-level API for matching query and reference sets, i.e., to compute pairwise similarity. Once the matching API is initialized with the identity database, one can simply feed it with images, and the matching API will return the most visually similar identity and appropriate image. For reference, see Figure\,\,\ref{fig:matching}.

\begin{figure}
    \centering
\begin{lstlisting}[language=Python,linewidth=1.0\linewidth]
import timm
import torchvision.transforms as T
from PIL import Image

# Load model from Huggingface Hub 
model = timm.create_model(
    "hf-hub:BVRA/MegaDescriptor-L-384",
    pretrained=True)

model = model.eval()

# Load expected image transformations
transforms = T.Compose([T.Resize(224), 
                        T.ToTensor(), 
                        T.Normalize(
                            [0.5, 0.5, 0.5], 
                            [0.5, 0.5, 0.5])]) 

# Load/feed-forward image to MegaDescriptor
image = Image.open("./test_image.png")

output = model(transforms(image).unsqueeze(0))
\end{lstlisting}
    \vspace{-0.25cm}
    \caption{\textbf{Inference with MegaDescriptor}. A code snippet showcasing inference with the pre-trained MegaDescriptor model.}
    \label{fig:feature_prediction}
    \vspace{-0.2cm}
\end{figure}

\begin{figure}
    \centering
\begin{lstlisting}[language=Python,linewidth=1.0\linewidth]
from wildlife_tools.data import FeatureDatabase
from wildlife_tools.inference import KnnMatcher

# Load database of image features
database = FeatureDatabase.from_file(
    "database_file"
)

# Extract features from query image
image = Image.open("./query_image.png")
query = model(transforms(image).unsqueeze(0))

# Find nearest match in database
matcher = KnnMatcher(database)
matcher([query])
>>> ["id_george"]
\end{lstlisting}
    \vspace{-0.25cm}
    \caption{\textbf{Matching with WildlifeDatasets}. A code snippet showcasing accessible matching with already loaded pre-trained model.}
    \label{fig:matching}
    \vspace{-0.1cm}
\end{figure}

\subsection{Community-driven extension}
Our toolkit is designed to be easily extendable, both in terms of functionality and datasets, and we welcome contributions from the community. In particular, we encourage researchers to contribute their datasets and methods to be included in the WildlifeDataset. The datasets could be used for the development of new methods and will become part of future versions of the MegaDescriptor, enabling its expansion and improvement. This collaborative approach aims to further drive progress in the application of machine learning in ecology.
Once introduced in communities such as \href{https://lila.science/}{LILA BC} or AI for Conversation Slack\footnote{With around 2000 members; experts on ecology and machine learning.}, the toolkit has a great potential to revolutionize the field.

\newpage

\section{MegaDescriptor -- Methodology}

Wildlife re-identification is usually formulated as a closed-set classification problem, where the task is to assign identities from a predetermined set of known identities to given unseen images. Our setting draws inspiration from real-life applications, where animal ecologists compare a reference image set (i.e., a database of known identities) with a query image set (i.e., newly acquired images) to determine the identities of the individuals in new images. 
In the search for the best suitable methods for the MegaDescriptor, we follow up on existing literature \cite{pedersen2022re,li2019atrw,deb2018face,miele2021revisiting} and focus on local descriptors and metric Learning.
We evaluate all the ablation studies over 29 datasets\footnote{We avoided Drosophila (low complexity and high image number)  and SeaTurleID2022 \cite{SeaTurtleID2022} due to its big overlap with SeaTurleIDHeads \cite{papafitsoros2022seaturtleid}.} provided through the WildlifeDataset toolkit.

\subsection{Local features approaches}
Drawing inspiration from the success of local descriptors in existing wildlife re-identification tools \cite{dunbar2021hotspotter, pedersen2022re}, we include the SIFT and Superpoint descriptors in our evaluation. The matching process includes the following steps: (i) we extract keypoints and their corresponding descriptors from all images in reference and query sets, (ii) we compute the descriptors distance between all possible pairs of reference and query images, (iii) we employ a ratio test with a threshold to eliminate potentially false matches, with the optimal threshold values determined by matching performance on the reference set, and (iv) we determine the identity based on the absolute number of correspondences, predicting the identity with highest number from reference set.

\subsection{Metric learning approaches}

Following the recent progress in human and vehicle \mbox{re-id}\cite{chen2017beyond,yan2021beyond,musgrave2020metric}, we select two metric learning methods for our ablation studies -- Arcface\,\cite{deng2019arcface} and Triplet loss\,\cite{schroff2015facenet} -- which both learn a representation function that maps objects into a deep embedding space. The distance in the embedded space should depend on visual similarity between all identities, i.e., samples of the same individual are close, and different identities are far away. CNN- or transformer-based architectures are usually used as feature extractors. 

The \textbf{Triplet loss} \cite{schroff2015facenet,hermans2017defense} involves training the model using triplets $(x_a, x_p, x_n)$, where the anchor $x_a$ shares the same label as the positive $x_p$, but a different label from the negative $x_n$. The loss learns embedding where the distance between $x_a$ and $x_p$ is small while distance between $x_a$ and $x_n$ is large such that the former pair should be distant to latter by at least a margin $m$.  Learning can be further improved by a suitable triplet selection strategy, which we consider as a hyperparameter. We consider 'all' to include all valid triplets in batch, 'hard' for triplets where $x_n$ is closer to the $x_a$ than the $x_p$ and 'semi' to select triplets where $x_n$ is further from the $x_a$ than the $x_p$.

The \textbf{ArcFace loss} \cite{deng2019arcface} enhances the standard softmax loss by introducing an angular margin $m$ to improve the discriminative capabilities of the learned embeddings. The embeddings are both normalized and scaled, which places them on a hypersphere with a radius of $s$. Value of scale $s$ is selected as hyperparameter.\vspace{-0.15cm} \\

\begin{figure*}[!t]
\centering
\vspace{-0.15cm}
\includegraphics[width=0.9\linewidth]{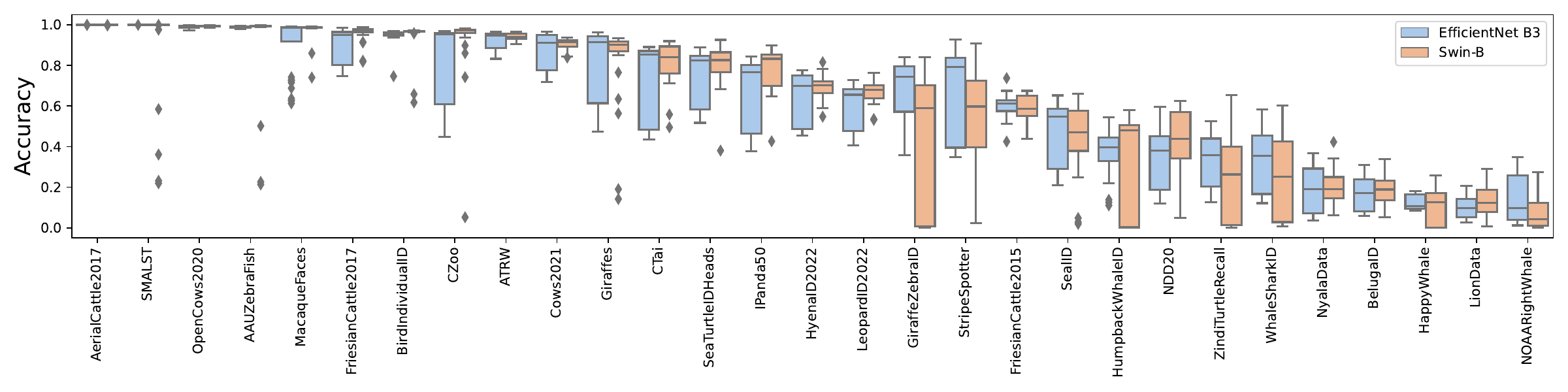}\vspace{-1.775cm}
\includegraphics[width=0.9\linewidth]{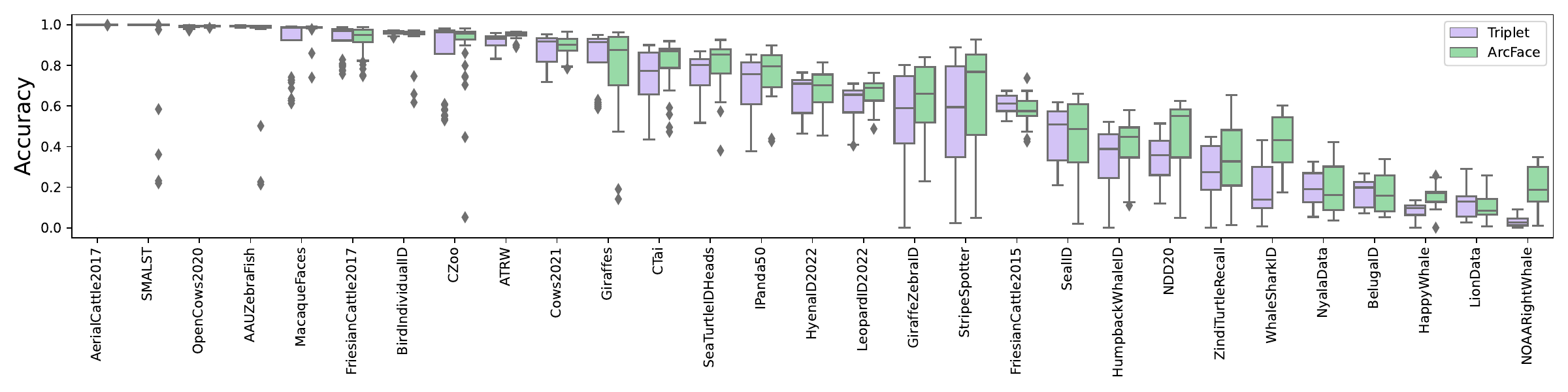}\vspace{-0.35cm}
\caption{\textbf{Ablation of the backbone architecture and metric learning method}. We compare two backbones -- Swin-B and EfficientNet-B3 -- and Triplet / ArcFace methods on all available animal re-id datasets. In most cases, the Swin-B with ArcFace maintains competitive or better performance than EfficientNet-B3 and Triplet.}
\label{fig:compare_architectures}
\end{figure*}

\noindent\textbf{Matching strategy}:  In the context of our extensive experimental scope, we adopt a simplified approach to determine the identity of query (i.e., test) images, relying solely on the closest match within the reference set. To frame this in machine learning terminology, we essentially create a 1-nearest-neighbor classifier within a deep-embedding space using cosine similarity. \vspace{-0.15cm} \\

\noindent\textbf{Training strategy}:
While training models, we use all 29 publicly available datasets provided through the WildlifeDataset toolkit. All datasets were split in an 80/20\% ratio for reference and query sets, respectively, while preserving the closed set setting, i.e., all identities in the query set are available in the reference set.  Models were optimized using the SGD optimizer with momentum (0.9) for 100 epochs using the cosine annealing learning rate schedule and mini-batch of 128. \vspace{-0.15cm} \\

\noindent\textbf{Hyperparameter tunning}: The performance of the metric learning approaches is usually highly dependent on training data and optimization hyperparameters\, \cite{musgrave2020metric}. Therefore, we perform an exhaustive hyperparameters search to determine optimal hyperparameters with sustainable performance in all potential scenarios and datasets for both methods. Besides, we compare two backbone architectures -- EfficientNet-B3 and Swin-B -- with a comparable number of parameters. We select EfficientNet-B3 as a representative of traditional convolutional-based and Swin-B as a novel transformer-based architecture. 

For each architecture type and metric learning approach, we run a grid search over selected hyperparameters and all the datasets. We consider 72 different settings for each dataset, yielding 2088 training runs. We use the same optimization strategy as described above. All relevant hyperparameters and their appropriate values are listed in Table\,\,\ref{table:hyperparameters}.

\begin{table}[!h]
\vspace{-0.2cm}
\small
\centering
\begin{tabular}{lc}
\toprule
\textbf{Backbone} & $\{\tt Swin-B, EfficientNet-B3\}$ \\
\textbf{Learning rate} & $\{\tt0.01, 0.001\}$ \\
\midrule
\textbf{ArcFace margin}~~~~ & $\{\tt0.25, 0.5, 0.75\}$ \\
\textbf{ArcFace scale} & $\{\tt32, 64, 128\}$ \\
\midrule
\textbf{Triplet mining} & $\{\tt all, semi, hard\}$ \\
\textbf{Triplet margin} & $\{\tt0.1, 0.2, 0.3\}$ \\
\bottomrule
\end{tabular}
\caption{\textbf{Grid-search setup}. Selected hyperparameters and their appropriate values for ArcFace and Triplet approaches.}
\label{table:hyperparameters}
\end{table}

\section{Ablation studies}

This section presents a set of ablation studies to empirically validate the design choices related to model distillation (i.e., selecting methods, architectures, and appropriate hyperparameters) while constructing the MegaDescriptor feature extractor, i.e., first-ever foundation model for animal re-identification. Furthermore, we provide both qualitative and quantitative performance evaluation comparing the newly proposed MegaDescriptor in a zero-shot setting with other methods, including SIFT, Superpoint, ImageNet, CLIP, and DINOv2.

\subsection{Loss and backbone components}
To determine the optimal metric learning loss function and backbone architecture configuration, we conducted an ablation study, comparing the performance (median accuracy) of ArcFace and Triplet loss with either a transformer- (Swin-B) or CNN-based backbone (EfficientNet-B3) on all available re-identification datasets. In most cases, the Swin-B with ArcFace combination maintains competitive or better performance than other variants. Overall, ArcFace and transformer-based backbone (Swin-B) performed better than Triplet and CNN backbone (EfficientNet-B3). First quantiles and top whiskers indicate that Triplet loss underperforms compared to ArcFace even with correctly set hyperparameters. The full comparison in the form of a box plot is provided in Figure\,\,\ref{fig:compare_architectures}.

\subsection{Hyperparameter tunning}
In order to overcome the performance sensitivity of metric learning approaches regarding hyperparameter selection and to select the generally optimal parameters,  we have performed a comprehensive grid search strategy.

Following the results from the previous ablation, we evaluate how various hyperparameter settings affect the performance of a Swin-B backbone optimized with Arcface and Triplet losses. 
In the case of ArcFace, the best setting (i.e., $lr=0.001$, $m=0.5$, and $s=64$) achieved a median performance of 87.3\% with 25\% and 75\% quantiles of 49.2\% and 96.4\%, respectively. Interestingly, three settings underperformed by a significant margin, most likely due to unexpected divergence in the training\footnote{These three settings were excluded from further evaluation and visualization for a more fair comparison.}. The worst settings achieved a mean accuracy of 6.4\%, 6.1\%, and 4.0\%. 
Compared to ArcFace, Triplet loss configurations showed higher performance on both 25\% and 75\% quantiles, indicating significant performance variability.

\begin{figure}[!ht]
\vspace{-0.15cm}
\centering
\includegraphics[width=0.985\linewidth]{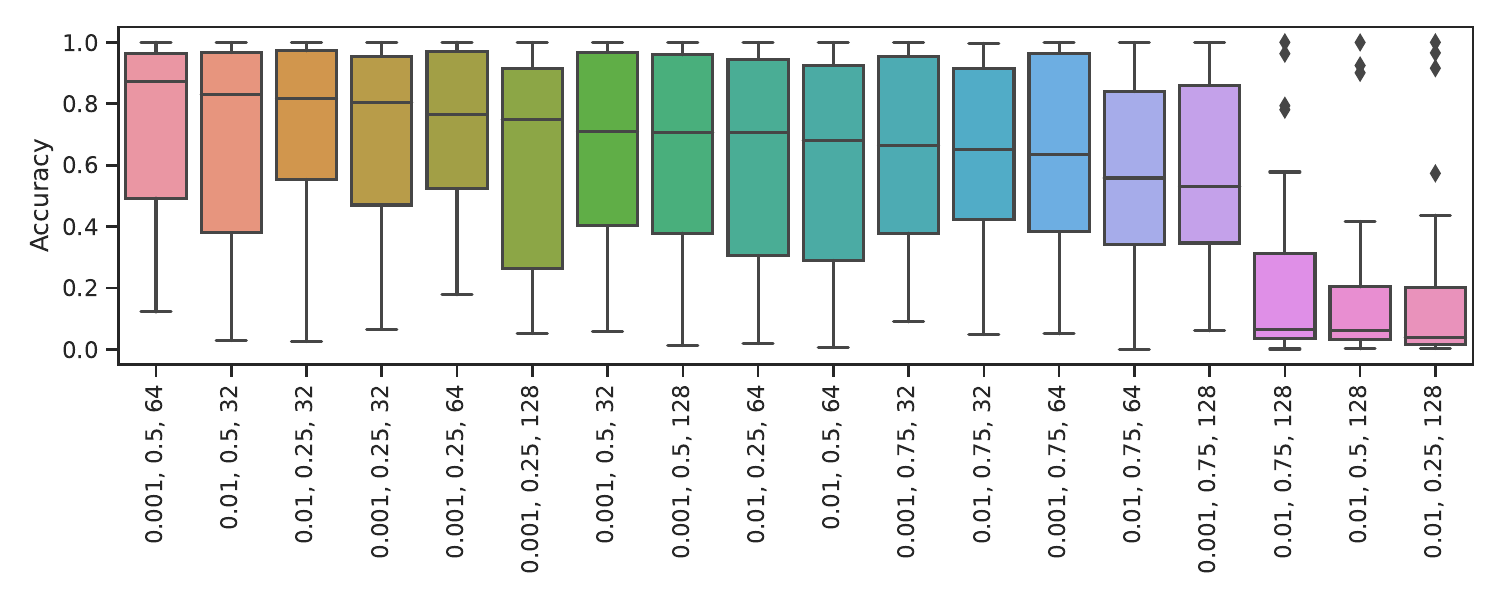}\vspace{-0.2cm}
\includegraphics[width=0.985\linewidth]{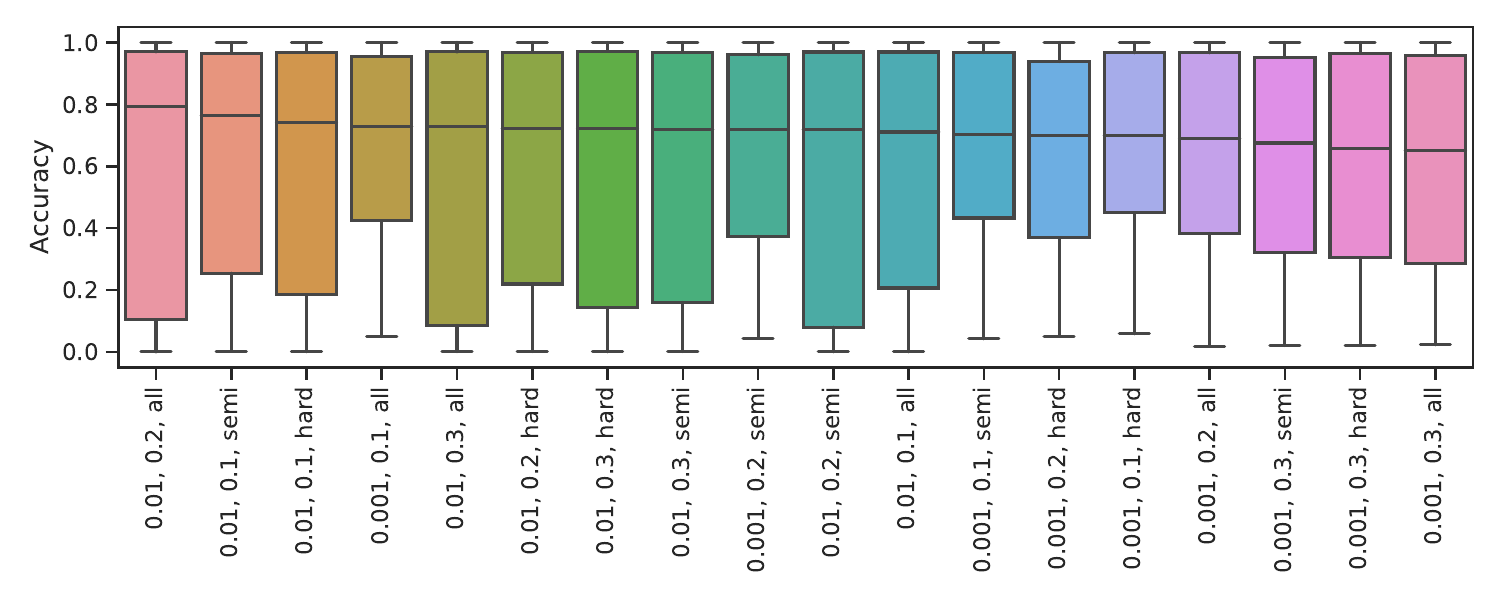}
\vspace{-0.2cm}
\caption{\textbf{Ablation of hyperparameters search}. We display performance for all settings as a boxplot combining accuracy from all 29 datasets. ArcFace (Top) and Triplet loss (Bottom).}
\label{fig:swin_tripler_hyperparameters}
\vspace{-0.2cm}
\end{figure}

The outcomes of the study are visualized in Figure \ref{fig:swin_tripler_hyperparameters} as a boxplot, where each box consists of 29 values.

\newpage

\subsection{Metric learning vs. Local features}

The results conducted over 29 datasets suggested that both metric learning approaches (Triplet and ArcFace) outperformed the local-feature-based methods on most datasets by a significant margin. The comparison of local-feature-based methods (SIFT and Superpoint) revealed that Superpoints are a better fit for animal re-identification, even though they are rarely used over SIFT descriptors in the literature. A detailed comparison is provided in Table\,\,\ref{table:metric_vs_local}. Note that the Giraffes dataset was labeled using local descriptors; hence, the performance is inflated and better than for metric learning.

The same experiment revealed that several datasets, e.g., AerialCattle2017, SMALST, MacaqueFaces, Giraffes, and AAUZebraFish, are solved or close to that point and should be omitted from development and benchmarking.

\begin{table}[!h]
\fboxsep=2.5pt
\def\arraystretch{0.25}
\setlength{\tabcolsep}{0.4em} 
\small
\centering
\begin{tabular}{@{}lcccc@{}}
\toprule
\footnotesize{\textbf{Dataset}}  & \footnotesize{\textbf{SIFT}} & \textbf{\footnotesize{Superpoint}} & \footnotesize{\textbf{Triplet}} & \footnotesize{\textbf{ArcFace}}  \\
\midrule
AAUZebraFish        & 65.09 & 25.09 & \colorbox{Greener}{99.40} & \colorbox{LightGreen}{98.95} \\
ATRW                & 89.30 & 92.74 & \colorbox{LightGreen}{93.26} & \colorbox{Greener}{95.63}  \\
AerialCattle2017    & 98.96 & 99.06 & \colorbox{Greener}{100.0} & \colorbox{Greener}{100.0}  \\
BelugaID            &~~1.10 &~~0.02 & \colorbox{Greener}{19.85} & \colorbox{LightGreen}{15.74}   \\
BirdIndividualID    & 48.96 & 48.71 & \colorbox{Greener}{96.45} & \colorbox{LightGreen}{96.00}  \\
CTai                & 33.87 & 29.58 & \colorbox{LightGreen}{77.44} & \colorbox{Greener}{87.14}  \\
CZoo                & 67.61 & 83.92 & \colorbox{Greener}{96.34} & \colorbox{LightGreen}{95.75}  \\
Cows2021            & 58.82 & 75.89 & \colorbox{Greener}{91.90} & \colorbox{LightGreen}{90.14}  \\
FriesianCattle2015  & 56.25 & 55.00 & \colorbox{Greener}{61.25} & \colorbox{LightGreen}{57.50}  \\
FriesianCattle2017  & 85.86 & 86.87 & \colorbox{Greener}{96.97} & \colorbox{LightGreen}{94.95}   \\
GiraffeZebraID      & \colorbox{Greener}{74.45} & \colorbox{LightGreen}{73.85} & 58.85 & 66.07   \\
Giraffes            & \colorbox{LightGreen}{97.01} & \colorbox{Greener}{99.25} & 91.42 & 88.69 \\
HappyWhale          &~~0.38 &~~0.42 &~~9.73 & \colorbox{Greener}{17.03}  \\
HumpbackWhaleID     & 11.65 & 11.82 & 38.78 & \colorbox{Greener}{44.75}  \\
HyenaID2022         & 39.84 & 46.67 & \colorbox{Greener}{71.03} & \colorbox{LightGreen}{70.32}  \\ 
IPanda50            & 35.12 & 47.35 & \colorbox{LightGreen}{75.71} & \colorbox{Greener}{79.71}  \\
LeopardID2022       & \colorbox{LightGreen}{72.71} & \colorbox{Greener}{75.08} & 65.56 & 69.02  \\
LionData            & \colorbox{Greener}{31.61} &~~5.16 & \colorbox{LightGreen}{12.90} &~~8.39  \\
MacaqueFaces        & 75.72 & 75.08 & \colorbox{LightGreen}{98.69} & \colorbox{Greener}{98.73}  \\
NDD20               & 17.14 & 29.01 & \colorbox{LightGreen}{35.88}& \colorbox{Greener}{55.18}   \\
NOAARightWhale      &~~6.53 & \colorbox{LightGreen}{15.31} &~~2.68 & \colorbox{Greener}{18.74}  \\
NyalaData           & 10.75 & 18.46 & \colorbox{LightGreen}{19.16} & \colorbox{Greener}{19.85}  \\
OpenCows2020        & 72.76 & 86.38 & \colorbox{LightGreen}{99.31} & \colorbox{Greener}{99.37}  \\
SMALST              & 92.22 & 98.37 & \colorbox{Greener}{100.0} & \colorbox{Greener}{100.0}  \\
SeaTurtleIDHeads    & 55.23 & \colorbox{LightGreen}{80.58} & 80.22 & \colorbox{Greener}{85.32}  \\
SealID              & 31.41 & \colorbox{Greener}{62.11} & \colorbox{LightGreen}{50.84} & 48.68  \\
StripeSpotter       & 70.12 & \colorbox{Greener}{94.51} & 59.45 & \colorbox{LightGreen}{76.83}  \\
WhaleSharkID        &~~4.29 & \colorbox{LightGreen}{22.90} & 13.88 & \colorbox{Greener}{43.10}  \\
ZindiTurtleRecall   & 17.91 & 25.73 & \colorbox{LightGreen}{27.40} & \colorbox{Greener}{32.74}  \\
\bottomrule
\end{tabular}
\caption{\textbf{Ablation of animal re-id methods}. We compare two local-feature (SIFT and Superpoint) methods with two metric learning approaches (Triplet and ArcFace). Metric learning approaches outperformed the local-feature methods on most datasets. ArcFace provides more consistent performance. For metric learning, we list the median from the previous ablation.}
\label{table:metric_vs_local}
\end{table}

\section{Performance evaluation}

Insights from our ablation studies led to the creation of MegaDescriptors -- the Swin-transformer-based models optimized with ArcFace loss and optimal hyperparameters using all publicly available animal re-id datasets.

In order to verify the expected outcomes, we perform a similar comparison as in metric learning vs. Local features ablation, and we compare the MegaDescriptor with CLIP (ViT-L/p14-336), ImageNet-1k (Swin-B/p4-w7-224), and DINOv2 (ViT-L/p14-518) pre-trained models. The proposed MegaDescriptor with Swin-L/p4-w12-384 backbone performs consistently on all datasets and outperforms all methods in on all 29 datasets. Notably, the state-of-the-art foundation model for almost any vision task -- DINOv2 -- with a much higher input size ($518\times518$) and larger backbone performs poorly in animal re-identification.

\begin{table}[!ht]
\fboxsep=2.5pt
\def\arraystretch{0.25}
\setlength{\tabcolsep}{0.2em} 
\small
\centering
\begin{tabular}{@{}lcccc@{}}
\toprule
\footnotesize{\textbf{Dataset}} & \footnotesize{\textbf{ImageNet}}  & \footnotesize{\textbf{CLIP}} & \footnotesize{\textbf{DINOv2}} & \footnotesize{\textbf{MegaDesc.}} \\
\midrule
AAUZebraFish       &     94.38 &     94.91 &  \colorbox{LightGreen}{96.93} & \colorbox{Greener}{99.93} \\
ATRW               &     88.37 &     86.88 &  \colorbox{LightGreen}{88.47} & \colorbox{Greener}{94.33} \\
AerialCattle2017   & \colorbox{Greener}{100.0} &  \colorbox{LightGreen}{99.99} &  \colorbox{Greener}{100.0} & \colorbox{Greener}{100.0} \\
BelugaID           & \colorbox{LightGreen}{19.58} &     11.20 &   14.64 & \colorbox{Greener}{66.48} \\
BirdIndividualID   &     63.11 &     52.75 & \colorbox{LightGreen}{74.90} & \colorbox{Greener}{97.82} \\
CTai               &     60.99 &     50.38 & \colorbox{LightGreen}{68.70} & \colorbox{Greener}{91.10} \\
CZoo               &     78.49 &     58.87 & \colorbox{LightGreen}{87.00} & \colorbox{Greener}{99.05} \\
Cows2021           &     57.84 &     41.06 & \colorbox{LightGreen}{58.19} & \colorbox{Greener}{99.54} \\
FriesianCattle2015 & \colorbox{Greener}{55.00} &   \colorbox{LightGreen}{53.75} &   \colorbox{Greener}{55.00} & \colorbox{Greener}{55.00} \\
FriesianCattle2017 & \colorbox{LightGreen}{83.84} &     79.29 &   80.30 & \colorbox{Greener}{96.46} \\
GiraffeZebraID     &     21.89 &     32.47 & \colorbox{LightGreen}{37.99} & \colorbox{Greener}{83.17} \\
Giraffes           &     59.70 &     42.16 & \colorbox{LightGreen}{60.82} & \colorbox{Greener}{91.04} \\
HappyWhale         &     14.25 & \colorbox{LightGreen}{15.30} &   13.26 & \colorbox{Greener}{34.30} \\
HumpbackWhaleID    &    \colorbox{LightGreen}{~~7.32} &    ~~3.23 &  ~~6.44 & \colorbox{Greener}{77.81} \\
HyenaID2022        &     46.83 &     45.71 & \colorbox{LightGreen}{49.52} & \colorbox{Greener}{78.41} \\
IPanda50           & \colorbox{LightGreen}{72.51} &     57.60 &   62.84 & \colorbox{Greener}{86.91} \\
LeopardID2022      & \colorbox{LightGreen}{61.13} &     59.94 &   57.50 & \colorbox{Greener}{75.58} \\
LionData           & \colorbox{LightGreen}{20.65} &    ~~5.16 &   12.90 & \colorbox{Greener}{25.16} \\
MacaqueFaces       &     78.58 &     64.17 & \colorbox{LightGreen}{91.56} & \colorbox{Greener}{99.04} \\
NDD20              &     43.13 & \colorbox{LightGreen}{46.70} &   37.85 & \colorbox{Greener}{67.42} \\
NOAARightWhale     & \colorbox{LightGreen}{28.37} &     28.27 &   24.84 & \colorbox{Greener}{40.26} \\
NyalaData          &     10.28 &     10.51 & \colorbox{LightGreen}{14.72} & \colorbox{Greener}{36.45} \\
OpenCows2020       & \colorbox{LightGreen}{92.29} &     82.26 &   90.18 & \colorbox{Greener}{100.0} \\
SMALST             &     91.25 &     83.04 & \colorbox{LightGreen}{94.63} & \colorbox{Greener}{100.0} \\
SeaTurtleIDHeads   &     43.84 &     33.57 & \colorbox{LightGreen}{46.08} & \colorbox{Greener}{91.18} \\
SealID             & \colorbox{LightGreen}{41.73} &     34.05 &   29.26 & \colorbox{Greener}{78.66} \\
StripeSpotter      &     73.17 &     66.46 & \colorbox{LightGreen}{82.93} & \colorbox{Greener}{98.17} \\
WhaleSharkID       & \colorbox{LightGreen}{28.26} &     26.37 &   22.02 & \colorbox{Greener}{62.02} \\
ZindiTurtleRecall  & \colorbox{LightGreen}{15.61} &     12.26 &   14.83 & \colorbox{Greener}{74.40} \\
\bottomrule
\end{tabular}
\caption{\textbf{Animal re-identification performance}. We compare the MegaDescriptor-L (Swin-L/p4-w12-384) among available pre-trained models, e.g., ImageNet-1k (Swin-B/p4-w7-224), CLIP (ViT-L/p14-336), and DINOv2 (ViT-L/p14-518). The proposed MegaDescriptor-L provides consistent performance on all datasets and outperforms all methods on all 29 datasets.}
\label{table:dataset_splits}
\end{table}

\begin{figure*}[!t]
\fboxsep=1.0pt
\vspace{-0.75cm}
\begin{center}
    \includegraphics[width=0.3\linewidth]{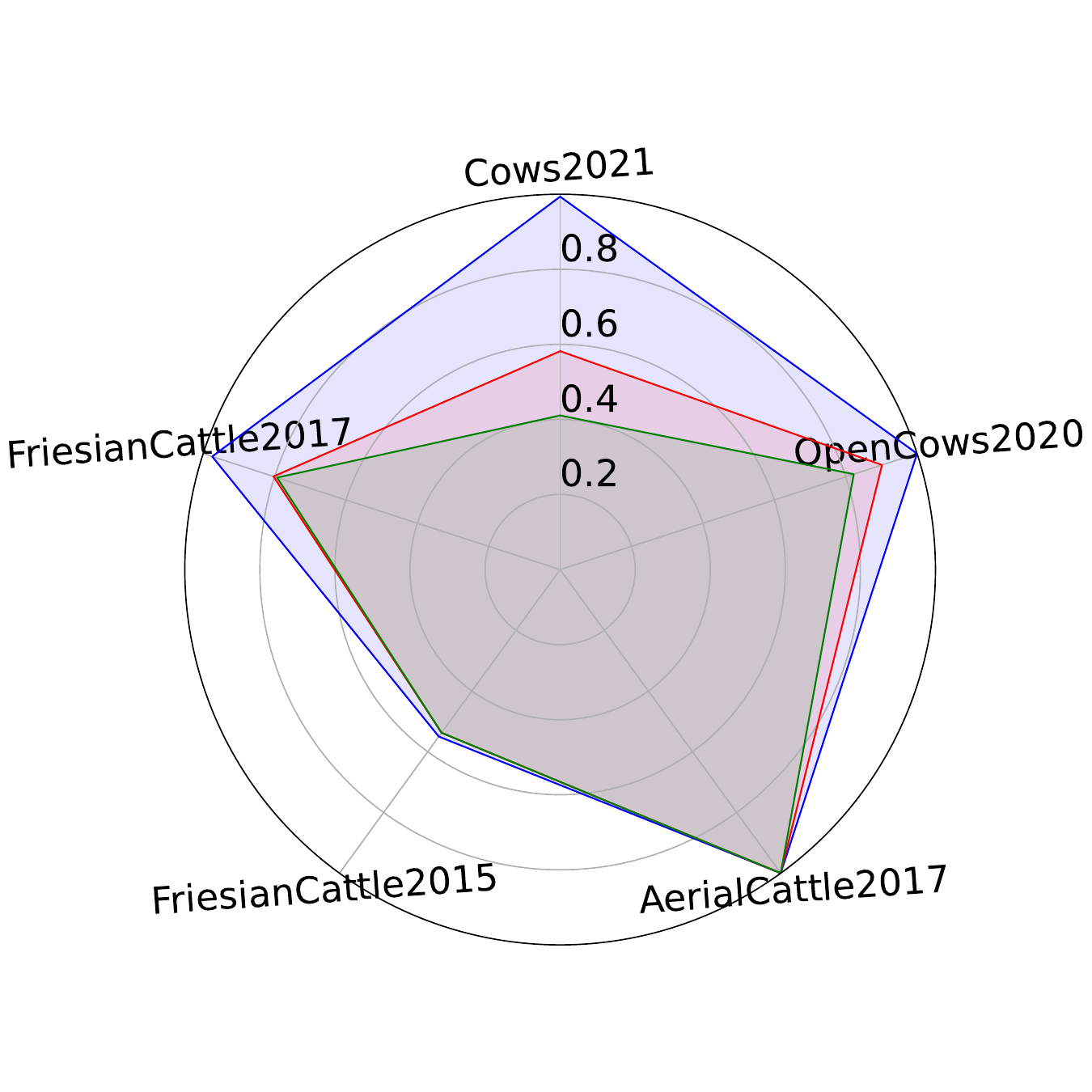}
    \includegraphics[width=0.3\linewidth]{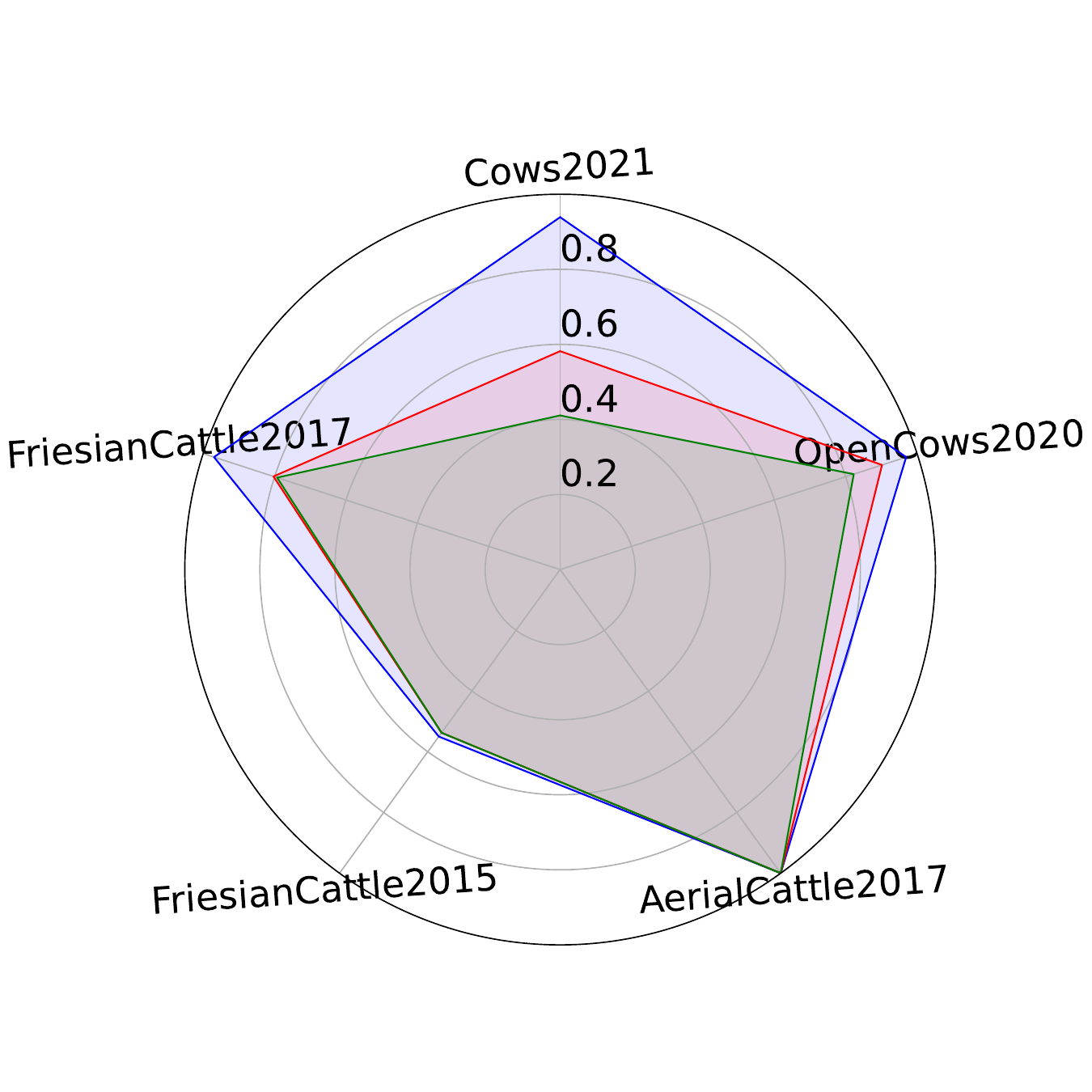}
    \includegraphics[width=0.3\linewidth]{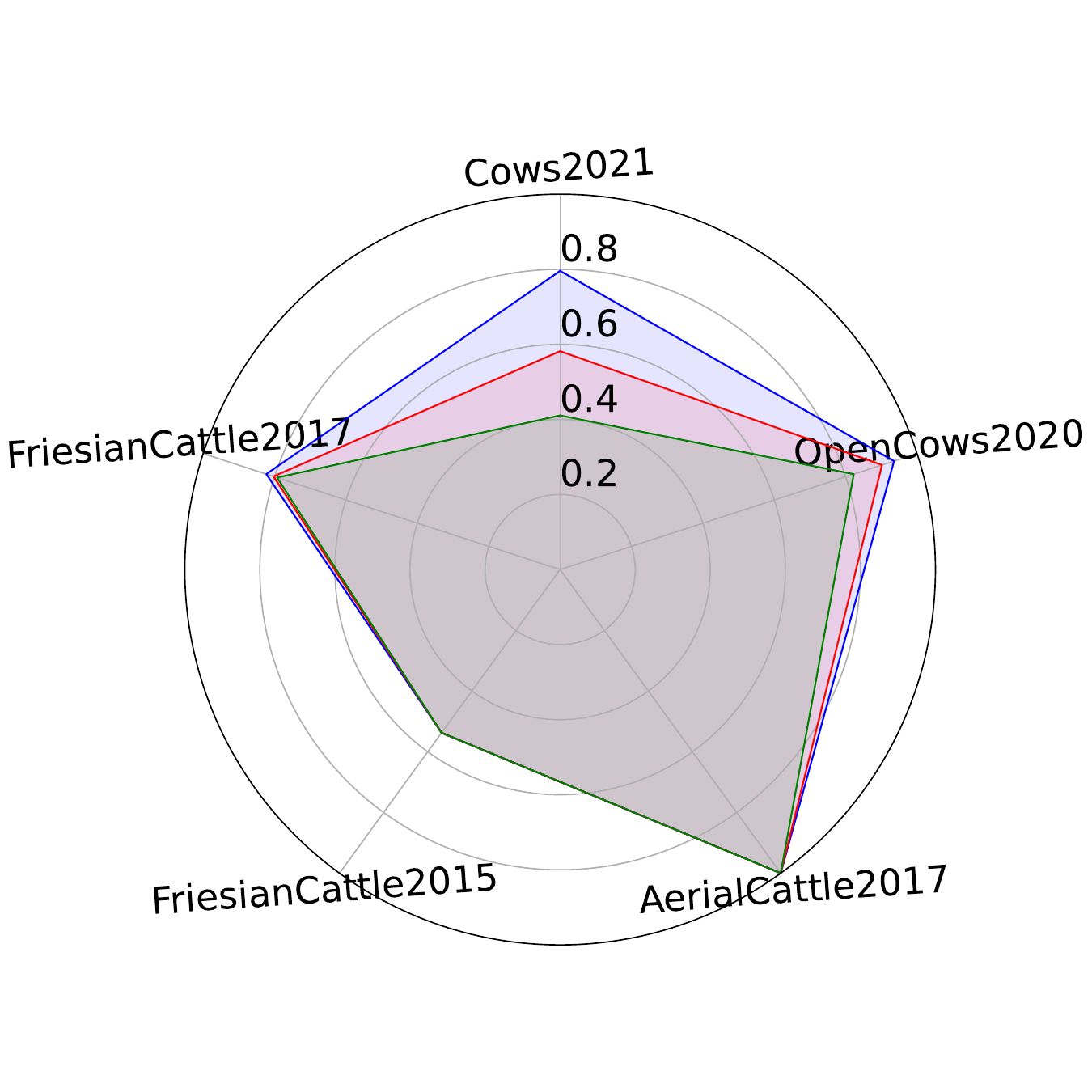}\vspace{-1.0cm}\\
\end{center}
\textit{\hspace{2.55cm}Same Dataset\hspace{3.4cm}Seen Domain\hspace{3.1cm}Unseen Domain}
\caption{\textbf{Seen domain and un-seen domain performance}. We compare the performance of a \colorbox{Mega}{MegaDescriptor-B} (Swin-B/p4-w7-224), \colorbox{ImageNet}{CLIP} (ViT-L/p14-336) and \colorbox{DINO}{DINOv2} (ViT-L/p14-518) on (i) \textit{Same Dataset}: all datasets were used for fine-tuning, (ii) \textit{Seen Domain}: Cows 2021 and OpenCows2020 were not used for fine-tuning, and (iii) \textit{Unseen Domain}: no datasets were used for fine-tuning.}
\label{fig:cattle-performance}
\end{figure*}

\begin{figure}[!b]
\vspace{-0.75cm}
\fboxsep=1.0pt
\centering
\includegraphics[width=0.9\linewidth]{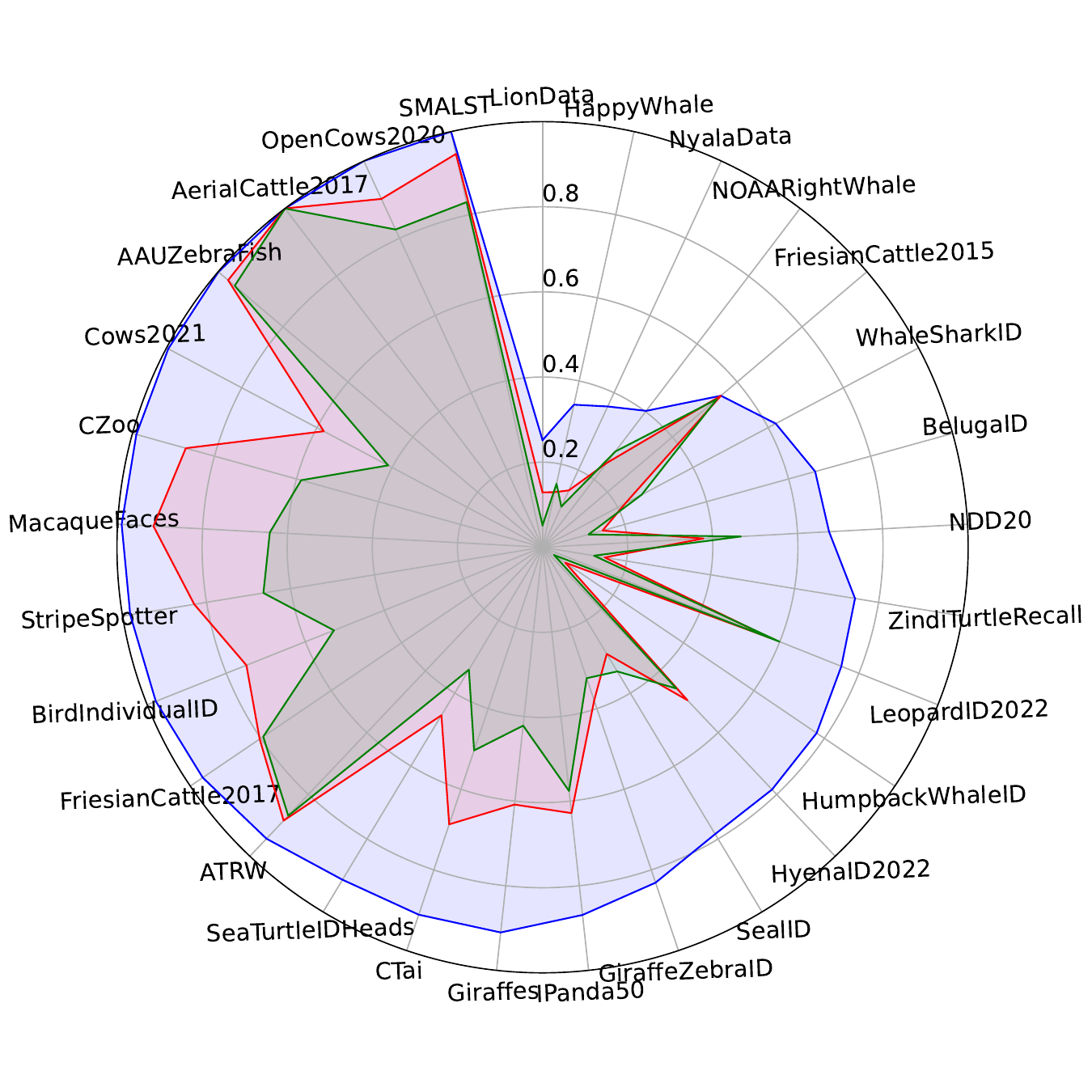}
\vspace{-0.5cm}
\caption{\textbf{\textit{Pre-trained} models performance evaluation.} We compare \colorbox{DINO}{DINOv2} (ViT-L/p14-518), 
\colorbox{ImageNet}{CLIP} (ViT-L/p14-336),
and \colorbox{Mega}{MegaDescriptor-L} (Swin-L/p4-w12-384) on 29 selected datasets.}
\label{fig:results_mega}
\end{figure}

\subsection{Seen and unseen domain performance}
This section illustrates how the proposed MegaDescriptor can effectively leverage features learned from different datasets and its ability to generalize beyond the datasets it was initially fine-tuned on. By performing this experiment, we try to mimic how the MegaDescriptor will perform on \textit{Seen (known)} and \textit{Unseen Domains (unknown)}.

We evaluate the generalization capabilities using the MegaDescriptor-B and all available datasets from one domain (cattle), e.g., AerialCattle2017, FriesianCattle2015, FriesianCattle2017, Cows2021, and OpenCows2020.
The first mutation (\textit{Same Dataset}) was trained on training data from all datasets and evaluated on test data. The second mutation (\textit{Seen Domain}) used just the part of the domain for training; OpenCows2020 and Cows2021 datasets were excluded. The third mutation (\textit{Unseen Domain}) excludes all the cattle datasets from training.

The MegaDescriptor-B, compared with a CLIP and DINOv2, yields significantly better or competitive performance (see Figure \ref{fig:cattle-performance}). This can be attributed to the capacity of MegaDescriptor to exploit not just cattle-specific features.
Upon excluding two cattle datasets (OpenCows2020 and Cows2021) from the training set, the MegaDescriptor's performance on those two datasets slightly decreases but still performs significantly better than DINOv2. 
The MegaDescriptor retains reasonable performance on the cattle datasets even when removing cattle images from training. We attribute this to learning general fine-grained features, which is essential for all the re-identification in any animal datasets, and subsequently transferring this knowledge to the re-identification of the cattle.
\section{Conclusion}

We have introduced the WildlifeDatasets toolkit, an open-source, user-friendly library that provides (i) convenient access and manipulation of all publicly available wildlife datasets for individual re-identification,
(ii) access to a variety of state-of-the-art models for animal re-identification, and (iii) simple API that allows inference and matching over new datasets. 
Besides, we have provided the most comprehensive experimental comparison of these datasets and essential methods in wildlife re-identification using local descriptors and deep learning approaches. 
Using insights from our ablation studies led to the creation of a MegaDescriptor, the first-ever foundation model for animal re-identification, which delivers state-of-the-art performance on a wide range of species. We anticipate that this toolkit will be widely used by both computer vision scientists and ecologists interested in wildlife re-identification and will significantly facilitate progress in this field.  \vspace{-0.35cm}

\blfootnote{This research was supported by the Czech Science Foundation (GA CR), project No. GA22-32620S and by the Technology Agency of the Czech Republic, project No. SS05010008. Computational resources were provided by the OP VVV project ``Research Center for Informatics'' (No. CZ.02.1.01/0.0/0.0/16\_019/0000765).}

\appendix
\section{WildlifeDatasets: Supplementary Materials}

\subsection{Ablation study on model size}
To showcase and quantify the performance of different MegaDescriptor flavors, we compare five variants, e.g., \textbf{B}ase, \textbf{S}mall, \textbf{T}iny, and \textbf{L}arge-224 and \textbf{L}arge-384, originating from corresponding variations of the Swin architecture. All the models were trained and evaluated using the same setting. 
Naturally, the model performance in terms of accuracy increased with an increasing model size, i.e., the MegaDescriptor-L-384 outperformed smaller flavors by a considerable margin in most cases. Overall, higher model complexity achieved higher performance with few exceptions, where it underperformed by a small margin, e.g., by 2.53\%, 0.48\%, and 0.08\% on FriesianCattle2017, LeopardID2022, and MacaqueFaces respectively. This is more or less statistically insignificant, given the poor quality of the data and the data acquisition. 

\begin{figure}[!b]
\vspace{-0.75cm}
    \fboxsep=1.0pt
    \centering
    \includegraphics[width=0.95\linewidth]{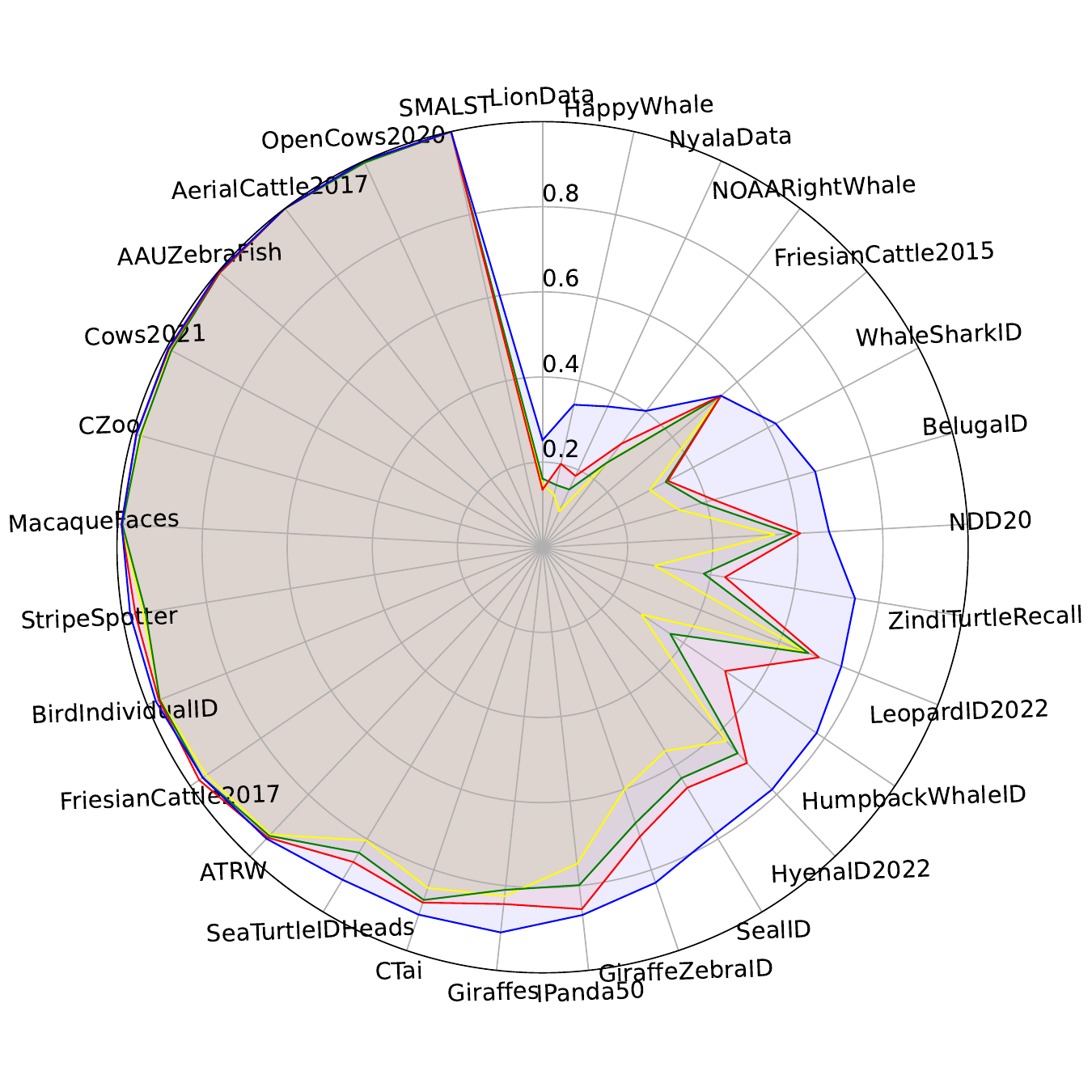}\vspace{-0.5cm}
    \caption{\textbf{Ablation study on model size/complexity.} Accuracy of different MegaDescriptor flavors \colorbox{Mega}{MegaDescriptor-L} (Swin-L/p4-w12-384), \colorbox{DINO}{MegaDescriptor-B} (Swin-B/p4-w7-224),
    \colorbox{ImageNet}{MegaDescriptor-S} (Swin-S/p4-w7-224),
    and \colorbox{Yellow}{MegaDescriptor-T} (Swin-T/p4-w7-224) on 29 animal re-identification datasets.}
    \label{fig:results_flavours}
\end{figure}

We visualized the accuracy of all provided MegaDescriptor flavors in Figure\,\,\ref{fig:results_flavours} and Table\,\,\ref{table:megadescriptors}.

\subsection{Online Documentation -- Dataset samples and tutorials}
We provide extensive \href{https://wildlifedatasets.github.io/wildlife-datasets/datasets/}{documentation} to give users a better orientation within the WildlifeDatasets toolkit and available features. It covers a wide range of use cases of the toolkit, including a guide to installation and dataset downloading, tutorials, and how to contribute. Notably, the documentation includes a detailed description of the datasets, including image samples.

\begin{table}[!ht]
\setlength{\tabcolsep}{0.3em}
\def\arraystretch{0.95}
\small
\centering
\begin{tabular}{@{}lccccc@{}}
\toprule
{} & \rotatebox{90}{\textbf{MegaDescriptor-T-224}} &  \rotatebox{90}{\textbf{MegaDescriptor-S-224}} & \rotatebox{90}{\textbf{MegaDescriptor-B-224}} &  \rotatebox{90}{\textbf{MegaDescriptor-L-224}} & \rotatebox{90}{\textbf{MegaDescriptor-L-384}}\\
\midrule
AAUZebraFish       &   99.55 &   99.55 &   99.63 & \colorbox{LightGreen}{99.85} & \colorbox{Greener}{99.93} \\
ATRW               &   93.02 &   93.40 & \colorbox{LightGreen}{93.95} &       93.67 & \colorbox{Greener}{94.33} \\
AerialCattle2017   & \colorbox{Greener}{100.0} & \colorbox{Greener}{100.0} & \colorbox{Greener}{100.0} & \colorbox{Greener}{100.0} & \colorbox{Greener}{100.0} \\
BelugaID           &   33.12 &   38.84 &   41.74 & \colorbox{LightGreen}{47.92} & \colorbox{Greener}{66.48} \\
BirdIndividualID   &   96.73 &   96.81 &   97.04 & \colorbox{LightGreen}{97.21} & \colorbox{Greener}{97.82} \\
CTai               &   84.46 &   87.46 &   88.10 & \colorbox{LightGreen}{90.68} & \colorbox{Greener}{91.10} \\
CZoo               &   97.87 &   98.11 & \colorbox{Greener}{99.05} & \colorbox{LightGreen}{98.35} & \colorbox{Greener}{99.05} \\
Cows2021           &   99.13 &   98.73 & \colorbox{LightGreen}{99.37} & \colorbox{LightGreen}{99.37} & \colorbox{Greener}{99.54} \\
FriesianCattle2015 & \colorbox{Greener}{55.00} & \colorbox{Greener}{55.00} & \colorbox{Greener}{55.00} & \colorbox{Greener}{55.00} & \colorbox{Greener}{55.00} \\
FriesianCattle2017 &   95.45 &   96.46 & \colorbox{LightGreen}{97.47} & \colorbox{Greener}{98.99} &       96.46 \\
GiraffeZebraID     &   60.15 &   68.40 &   71.72 & \colorbox{LightGreen}{78.04} & \colorbox{Greener}{83.17} \\
Giraffes           &   82.46 &   80.97 &   84.33 & \colorbox{LightGreen}{87.69} & \colorbox{Greener}{91.04} \\
HappyWhale         &   12.58 &   14.98 &   20.07 & \colorbox{LightGreen}{25.34} & \colorbox{Greener}{34.30} \\
HumpbackWhaleID    &   28.12 &   36.25 &   51.83 & \colorbox{LightGreen}{63.54} & \colorbox{Greener}{77.81} \\
HyenaID2022        &   62.70 &   66.67 &   69.84 & \colorbox{LightGreen}{77.30} & \colorbox{Greener}{78.41} \\
IPanda50           &   74.84 &   79.85 & \colorbox{LightGreen}{85.53} & 85.45 & \colorbox{Greener}{86.91} \\
LeopardID2022      &   67.06 &   67.27 &   69.92 & \colorbox{Greener}{76.06} & \colorbox{LightGreen}{75.58} \\
LionData           &   14.84 &   16.13 &   13.55 & \colorbox{LightGreen}{20.65} & \colorbox{Greener}{25.16} \\
MacaqueFaces       & \colorbox{LightGreen}{99.04} &   98.89 & \colorbox{Greener}{99.12} &       98.96 & \colorbox{LightGreen}{99.04} \\
NDD20              &   54.61 &   58.57 &   60.64 & \colorbox{LightGreen}{61.58} & \colorbox{Greener}{67.42} \\
NOAARightWhale     &   25.16 &   24.95 &   30.51 & \colorbox{LightGreen}{34.69} & \colorbox{Greener}{40.26} \\
NyalaData          & ~~9.35 &   14.95 &   18.46 & \colorbox{LightGreen}{21.73} & \colorbox{Greener}{36.45} \\
OpenCows2020       &   99.58 &   99.58 & \colorbox{Greener}{100.0} & \colorbox{LightGreen}{99.79} & \colorbox{Greener}{100.0} \\
SMALST             & \colorbox{Greener}{100.0} & \colorbox{Greener}{100.0} & \colorbox{Greener}{100.0} & \colorbox{Greener}{100.0} & \colorbox{Greener}{100.0} \\
SeaTurtleIDHeads   &   80.38 &   83.74 &   86.31 & \colorbox{LightGreen}{89.86} & \colorbox{Greener}{91.18} \\
SealID             &   55.88 &   63.31 &   65.95 & \colorbox{LightGreen}{70.02} & \colorbox{Greener}{78.66} \\
StripeSpotter      &   95.12 &   94.51 &   96.95 & \colorbox{LightGreen}{97.56} & \colorbox{Greener}{98.17} \\
WhaleSharkID       &   28.58 &   32.74 &   33.31 & \colorbox{LightGreen}{50.03} & \colorbox{Greener}{62.02} \\
ZindiTurtleRecall  &   26.77 &   38.38 &   43.45 & \colorbox{LightGreen}{58.14} & \colorbox{Greener}{74.40} \\
\bottomrule
\end{tabular}
\caption{\textbf{Ablation study on model size/complexity}. We compare five MegaDescriptor flavors, e.g., \textbf{L}arge, \textbf{B}ase, \textbf{S}mall, and \textbf{T}iny, in terms of accuracy.
In general, models with a bigger model size or higher input resolution outperformed their \textit{smaller} variants by a considerable margin. The best-performing model -- MegaDescriptor-L-384 -- underperformed by 2.53\%, 0.48\%, and 0.08\% on FriesianCattle2017, LeopardID2022, and MacaqueFaces, respectively.
}
\vspace{2cm}
\label{table:megadescriptors}
\end{table}

{\small
\bibliographystyle{ieee_fullname}
\bibliography{references,sn-bibliography}
}

\end{document}